\documentclass{article}
\usepackage{arxiv}
\usepackage[utf8]{inputenc} 
\usepackage[T1]{fontenc}    
\usepackage[hidelinks]{hyperref}       
\usepackage{url}            
\usepackage{booktabs} 
\usepackage{siunitx}        
\usepackage{amsfonts}       
\usepackage{nicefrac}       
\usepackage{microtype}    

\usepackage{algorithm}
\usepackage{algpseudocode}
\usepackage{amssymb,amsbsy, amsmath}
\usepackage[dvipsnames,usenames]{color}
\usepackage{graphicx}
\usepackage{rotating}
\usepackage[export]{adjustbox}

\usepackage{dcolumn}
\usepackage[normalem]{ulem} 
\usepackage{xr}
\usepackage{nameref}
\usepackage[version=4]{mhchem}
\usepackage[square, comma, numbers, sort&compress]{natbib}
\definecolor{cadmiumgreen}{rgb}{0.0, 0.42, 0.24}
\usepackage{authblk}
\usepackage{zref-xr}
\usepackage{booktabs} 
\usepackage{longtable} 
\usepackage[framemethod=TikZ]{mdframed}
\usepackage{arydshln}
\definecolor{customcolor}{HTML}{fffdf0}
\usepackage{tabularx}
\usepackage{authblk}

\zxrsetup{toltxlabel}

\usepackage{setspace}
\usepackage{fullwidth}

\begin{document}

\title{Agent-based Learning of materials\\ datasets from scientific literature}

\author[1,2]{Mehrad Ansari}
\author[ \space ,1,2]{Seyed Mohamad Moosavi\thanks{Corresponding author: mohamad.moosavi@utoronto.ca}}

\affil[1]{Acceleration Consortium, University of Toronto, Toronto, Ontario M5S 3E5, Canada}
\affil[2]{ Department of Chemical Engineering \& Applied Chemistry, University of Toronto, Toronto, Ontario M5S 3E5, Canada}

\maketitle
\begin{abstract}
Advancements in machine learning and artificial intelligence are transforming materials discovery. Yet, the availability of structured experimental data remains a bottleneck. 
The vast corpus of scientific literature presents a valuable and rich resource of such data. 
However, manual dataset creation from these resources is challenging due to issues in maintaining quality and consistency, scalability limitations, and the risk of human error and bias. Therefore, in this work, we develop a chemist AI agent, powered by large language models (LLMs), to overcome these challenges by autonomously creating structured datasets from natural language text, ranging from sentences and paragraphs to extensive scientific research articles.
Our chemist AI agent, \emph{Eunomia}, can plan and execute actions by leveraging the existing knowledge from decades of scientific research articles, scientists, the Internet and other tools altogether.
We benchmark the performance of our approach in three different information extraction tasks with various levels of complexity, including solid-state impurity doping, metal-organic framework (MOF) chemical formula, and property relations.
Our results demonstrate that our zero-shot agent, with the appropriate tools, is capable of attaining performance that is either superior or comparable to the state-of-the-art fine-tuned materials information extraction methods.
This approach simplifies compilation of machine learning-ready datasets for various materials discovery applications, and significantly ease the accessibility of advanced natural language processing tools for novice users in natural language.
The methodology in this work is developed as an open-source software on \href{https://github.com/AI4ChemS/Eunomia}{\color{blue}{{https://github.com/AI4ChemS/Eunomia}}}.

\end{abstract}
\section{Introduction}

The past decade's extraordinary achievements in leveraging machine learning for chemical discovery highlight the power of accessible knowledge and structured data~\cite{moosavi2020role,butler2018machine,sanchez2018inverse}. 
However, a significant portion of chemical knowledge, particularly the experimental ones, is scattered across scientific literature in an unstructured format\cite{sayeed2023quantifying}. 
Researchers face challenges in effectively utilizing existing knowledge for design of experiments, as well as in comprehending the entirety of previous works in a field.
Thus, the development of methodologies to extract information from the literature and convert it into structured data will play a fundamental role in advancing the machine learning for molecules and materials.

Natural Language Processing (NLP) is a powerful tool for extracting information from scientific literature. Conventional NLP methods have been used in materials and chemical sciences~\cite{weston2019named,swain2016chemdataextractor, dunn2022structured, nandy2021using, kim2017materials, glasby2023digimof} for Named Entity Recognition. However, these methods are limited in other NLP tasks that are needed for a general-purpose data extraction tool, including Co-reference Resolution, Relation Extraction, Template Filling, Argument Mining, and Entity Linking. To better understand these NLP terminologies, let us consider an example taken from an abstract of a materials paper~\cite{li2013systematic} in the field of metal-organic frameworks (MOFs):

\begin{mdframed}[leftmargin=0.06\textwidth,
  rightmargin=0.06\textwidth,
  linewidth=0.8pt,
  innerleftmargin=10pt, 
  innerrightmargin=10pt, 
  innertopmargin=10pt, 
  innerbottommargin=10pt,
  backgroundcolor=customcolor]
An isoreticular series of cobalt-adeninate bio-MOFs (\ce{bio-MOFs}-11--14) is reported. The pores of \ce{bio-MOFs}-11--14 are decorated with acetate, propionate, butyrate, and valerate, respectively. The nitrogen (\ce{N2}) and carbon dioxide (\ce{CO2}) adsorption properties of these materials are studied and compared. The isosteric heats of adsorption for \ce{CO2} are calculated, and the \ce{CO2} : \ce{N2} selectivities for each material are determined. As the lengths of the aliphatic chains decorating the pores in \ce{bio-MOFs}-11--14 increase, the BET surface areas decrease from 1148 m\textsuperscript{2} g\textsuperscript{-1} to 17 m\textsuperscript{2} g\textsuperscript{-1} while the \ce{CO2} : \ce{N2} selectivities predicted from ideal adsorbed solution theory at 1 bar and 273 K for a 10 : 90 \ce{CO2} : \ce{N2} mixture range from 73 : 1 for \ce{bio-MOF}-11 to 123 : 1 for \ce{bio-MOF}-12 and finally to 107 : 1 for \ce{bio-MOF}-13. At 298 K, the selectivities are 43 : 1 for \ce{bio-MOF}-11, 52 : 1 for \ce{bio-MOF}-12, and 40 : 1 for \ce{bio-MOF}-13.
The water stability of \ce{bio-MOFs}-11--14 increases with increasing aliphatic chain length.
\end{mdframed}

\begin{itemize}
    \item \textbf{Named Entity Recognition}
    involves identifying and classifying the specific entities within the text into predefined categories (i.e., \emph{chemical compounds}: ``bio-MOFs-11–14'', ``acetate'', \emph{experimental conditions}: ``1 bar'', ``273 K'', ``10 : 90 CO2 : N2 mixture'').
    \item \textbf{Co-reference Resolution} focuses on finding all expressions that refer to the same entity in the text.
    As an example, phrases like ``these materials'', ``each material'' are references that relate back to the bio-MOFs-11-14 mentioned in the first sentence.
    \item \textbf{Relation Extraction} involves extracting semantic relationships from the text, which usually occur between two or more entities (i.e., the impact of ``aliphatic chain lengths'' on ``BET surface areas'' and ``CO2 : N2 selectivities'').
    \item \textbf{Template Filing}
    is an efficient approach to extract and structure complex information from text.
    As an example: \emph{material name}: bio-MOFs-11–14.
    \item \textbf{Argument Mining} focuses on the automatic identification and  extracts the reasoning presented within the text.
    As an example, the ``increase in the water stability'' of the mentioned MOFs are connected to the ``increasing length of the aliphatic chains''.
    \item \textbf{Entity Linking} takes
    one step further than named entity recognition and distinguishes between similarly named entities (i.e., the term ``bio-MOFs'' would be linked to databases or literature that describe these materials in detail.
\end{itemize}

The emergence of Large Language Models (LLMs) or foundation models, shows a great promise in tackling these complex NLP tasks \cite{dunn2022structured,polak2023extracting,zheng2023chatgpt,zheng2023image}. 
\citet{huang2022batterybert} fine-tuned a language model (BERT) on battery publications to extract device-level information from a paragraph that contains one device only. ~\citet{dunn2022structured} showed that fine-tuned LLMs using 100-1000 data points can perform Relation Extraction as well as template filling, enabling conversion of the extracted information into user-defined output formats. Despite these promising results, these methods require training data, limiting their ease of use and broad applicability. Moreover, LLM based approaches have not been explored for more intricate challenges, such as argument mining and co-reference resolution. These tasks are critical for practically using NLP for automated database development. For example, in one article, multiple materials might be discussed and authors use abbreviations like ``compound 1'' or simply ``1'' in the entire research manuscript for referencing after initially defining the chemical compound in the introduction section. Additionally, description of material properties often come with various interpretations, limiting using rigid name entity matching. As implementations of standalone LLMs fall short in addressing these intricate tasks, new methods are needed to enable reliable information extraction. An effective approach is to augment LLMs with domain-specific toolkits. These specialized tools offer precise answers, thus addressing the inherent limitations of LLMs in specific domains, and enhancing their overall performance and applicability~\cite{white2023assessment, bran2023chemcrow,boiko2023emergent, lala2023paperqa}.

\begin{figure}[ht]
    \centering
    \includegraphics[width=0.96\textwidth]{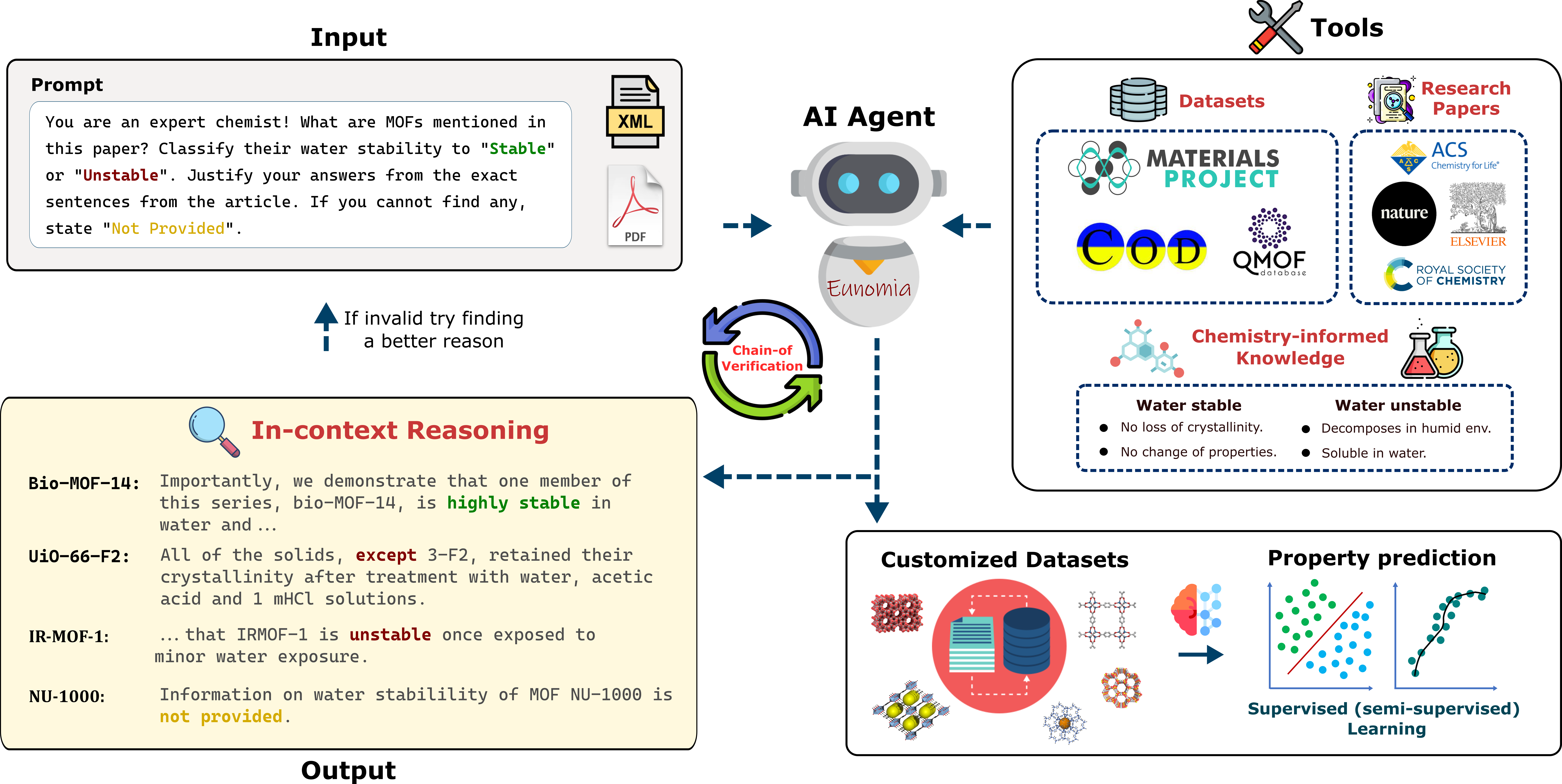}
    \caption{\textbf{Agent-based learning framework overview.}
    The AI agent equipped with various tools (online dataset search, document search, etc.) is tasked to extract information. The example shows the task of identifying all MOFs from a given research article, and predicting their property (e.g. water stability) by providing the reasoning for its decision.
    This reasoning is the exact in-context sentence from the paper, which is autonomously re-evaluated via the chain-of-verification tool of the agent to ensure its actual logical connection to the water stability property and reduce likelihood of hallucinations.
    The agent outputs a customized dataset that can be used to develop supervised or unsupervised machine learning methods. 
    \label{fig:toc}}
\end{figure}

In this work, we introduce an autonomous AI agent, \emph{Eunomia},
augmented with chemistry-informed tools, designed to extract materials science-relevant information from unstructured text and convert it into structured databases.
With an LLM at its core, our AI agent is capable of strategizing and executing actions by tapping into a wealth of knowledge from academic publications, domain-specific experts, the Internet, and other user-defined resources (see Figure~\ref{fig:toc}).
We show that this method streamlines data extraction, achieving remarkable accuracy and performance solely with a pre-trained LLM (GPT-4~\cite{openai2023gpt4}), eliminating the need for fine-tuning.
It offers adaptability by accommodating a variety of information extraction tasks through natural language text prompts for new output schemas and reducing the risk of hallucinations through a chain-of-verification process. 
This capability extends beyond what a standalone LLM can offer. \emph{Eunomia} simplifies the development of tailored datasets from literature for domain experts, eliminating the need for extensive programming, NLP, or machine learning expertise.

This manuscript is organized as follows;
Benchmarking and evaluating the model performance on three different materials NLP tasks with varying level of complexity are represented in Section~\ref{sec:case-studies}. This is followed by Section~\ref{sec:discuss}, with a discussion on the implications of our findings, the advantages and limitations of our approach, as well as suggested directions for future work.
Finally, in Section~\ref{sec:methods}, we describe our methodology on agent's toolkits and evaluation metrics.

\label{sec:results}
\section{AI Agent}
\label{sec:AI-agent}

In the realm of artificial intelligence, an ``agent'' is an autonomous entity capable of taking action based on its environment. In this work, we developed a chemist AI agent, \emph{Eunomia}, to autonomously extract information from scientific literature (Figure~\ref{fig:toc}). We use an LLM to serve as the brain of our agent~\cite{kojima2022large}. The LLM is equipped with advanced capabilities like planning and tool use to act beyond just a text generator, and act as a comprehensive problem solver, enabling effective interactions with the environment. We use ReAct architecture~\cite{yao2022react} for planning, enabling both reasoning and action. Our agent can interact with external sources like knowledge bases or environments to obtain more information. These knowledge bases are developed as toolkits (see method section for details) allowing the agent to extract relevant information from research articles, publicly available datasets, and built-in domain-specific chemical knowledge, ensuring its proficiency in playing the role of an expert chemist. We use OpenAI's GPT-4~\cite{openai2023gpt4} with a temperature of zero as our LLM and  LangChain~\cite{chase2022langchain} for the application framework development (note the choice of LLM is only a hyperparameter and other LLMs can be also used with our agent).
The application of LLMs in developing autonomous agents is a growing area of research~\cite{bran2023chemcrow,nakano2021webgpt,yao2022react, schick2023toolformer, lu2023chameleon, qin2023tool}, with a detailed survey available in Ref.~\cite{xi2023rise} for further insights.

\begin{figure}[ht!]
    \centering
    \includegraphics[width=\textwidth]{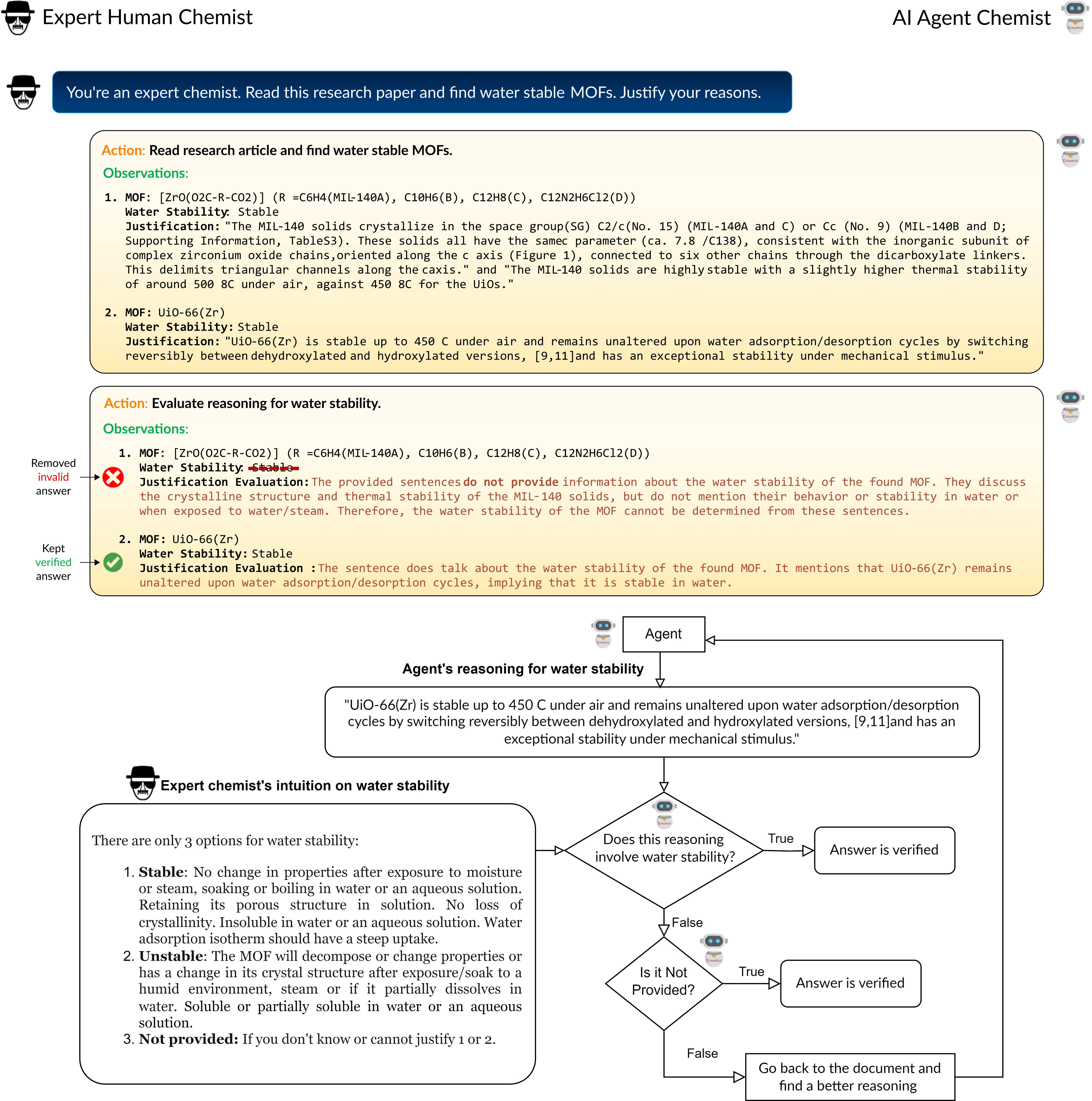}
    \caption{\textbf{Iterative Chain of Verification (CoV)}. 
    The agent is tasked with reading a materials research article and predicting water stability of any mentioned MOFs by providing reasoning. 
    In the initial run, the agent confuses water stability with thermal stability and  mistakenly predicts the second MOF as water-stable.
    The CoV tool evaluates the agent's decisions in its precious step by validating the justification against the pre-defined water stability criteria and disregards this prediction.
    \label{fig:CoV}}
\end{figure}

In addition to the standard search and text manipulation tools, we have implemented a Chain-of-Verification (CoV) tool to enhance the robustness of our AI agent against hallucination. Hallucination in a LLM refers to the generation of content that strays from factual reality or includes fabricated information~\cite{rawte2023survey}. In the CoV approach, the agent iteratively assesses its responses to ensure they remain logically consistent and coherent (see method section~\ref{sec:CoV} for details). This addition helps particularly with eliminating mistakenly extracted data related to semantically similar properties. An illustrative example is the case of stability of materials, where thermal, mechanical, and chemical stabilities might be confused by the agent. Figure~\ref{fig:CoV} illustrates how CoV process works in action: the agent is tasked to identify MOFs and the corresponding water stability data in a paper. The agent initially misclassifies a thermally stable MOF as water-stable, but then it corrects this mistake by a comprehensive review using the CoV tool. This tool improves the performance of the agent and ensures robust data extraction. 

\section{Case studies}
\label{sec:case-studies}
We evaluate the performance of our AI agent by benchmarking it across three different materials NLP tasks, with increasing task complexity (Table~\ref{tab:case_studies_overview}). In our assessment, we include a wide range of text lengths, including sentences, paragraphs and entire manuscript, as well as different NLP tasks discussed in the introduction. 
The first case study focuses on assessing the agent's performance on NLP tasks of lower complexity, specifically Named Entity Recognition and Relation Extraction.
For this, we use our agent to extract host-to-many dopants relationships from a single sentence. 
The second case study, with medium NLP complexity, involves obtaining MOFs' chemical formula and their corresponding guest species from a paragraph with multiple sentences. 
Finally, the third case study centers on predicting a given property of MOFs based on the context coming from a materials research paper. 
The property of interest in our work is water stability.
This case study aims to, in addition to Named Entity Recognition and Relation Extraction, evaluate the co-reference resolution and argument mining proficiency of our AI agent, tailored for chemists, which involves a high level of NLP complexity.
In all case studies, our chemist AI agent, Eunomia, is a zero-shot learner that is equipped with the Doc Search tool (see Section~\ref{sec:doc_search}). 
We have also conducted additional experiments by equipping Eunomia with the chain-of-verification (CoV) tool, as described in Section~\ref{sec:CoV}.
This is referred to as Eunomia + CoV from here on.

To fairly compare the performance of our agent with the state-of-the-art fine-tuned LLM methods, the evaluation methodology for the first two case studies mirrors precisely that of \citet{dunn2022structured} (see Section~\ref{sec:eval-metrics} for details), serving as a benchmark reference. In this study, ~\citet{dunn2022structured} is referred to as LLM-NERRE, which involves fine-tuning a pre-trained LLM (GPT-3~\cite{brown2020language}) on 100-1000 set of manually annotated examples, and then using the model to extract information from unstructured text.

\begin{table}[h!]
    \centering
    \caption{Overview of the three case studies based on their context from which data is extracted, NLP tasks and complexity}
    \small
    \begin{tabular}{l l p{4cm} c} 
        \toprule
        {Case Study} & {Context} & {NLP Tasks} & {Task Complexity} \\
        \midrule
        1.  Host-to-many Dopants Relation & Sentence & Named Entity Recognition, Relation Extraction & \includegraphics[height=1.45em,valign=c]{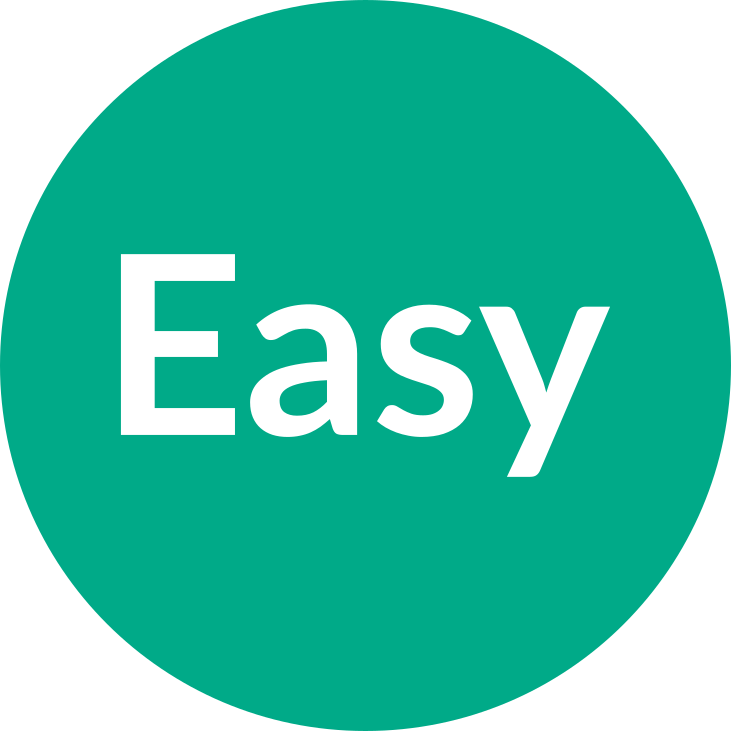}\vspace{1mm}\\ 

        2.   MOF Formula and Guest Species Relation & Paragraph & Named Entity Recognition, Relation Extraction & \includegraphics[height=1.45em,valign=c]{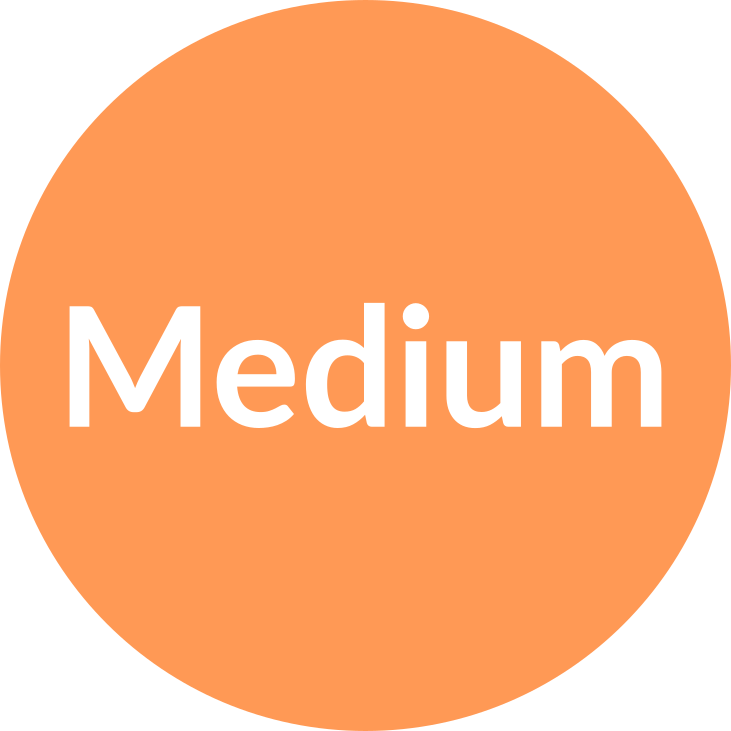}\vspace{1mm}\\ 

        3. MOF Property Relation & Research paper & Named Entity Recognition, Relation Extraction, Template Filing, Argument Mining, Entity Linking & \includegraphics[height=1.45em,valign=c]{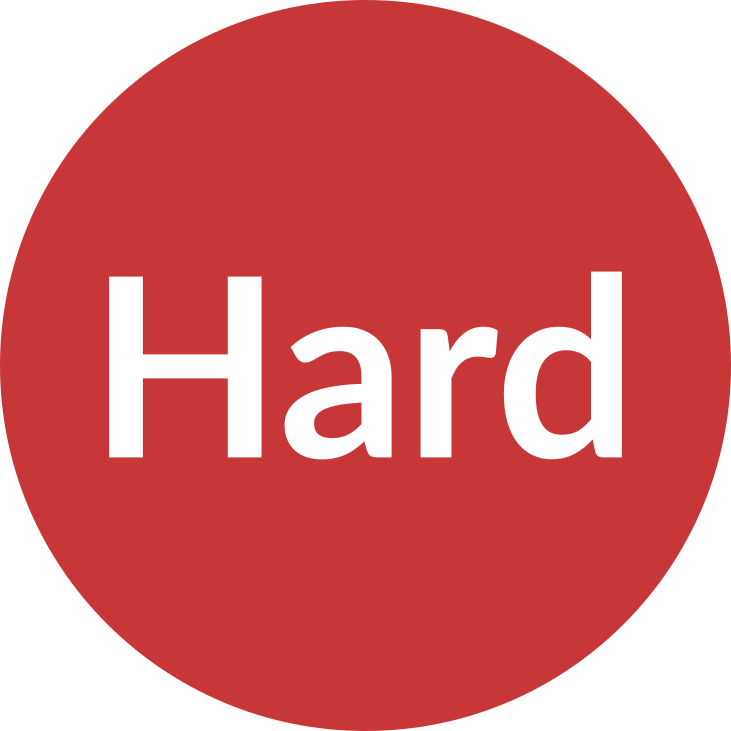}\vspace{1mm}\\ 
    \end{tabular}
    \label{tab:case_studies_overview}
\end{table}

\subsection{Case Study 1: Host-to-many Dopants Relation 
\includegraphics[height=1.45em,valign=c]{figures/easy.png}\vspace{1mm}}
\label{sec:dopant}

This case study aims to extract structured information about solid-state impurity doping from a single sentence. 
The objective is to identify the two entities ``host'' and ``dopant''. ``Host'' refers to the foundational crystal, sample, or category of material characterized by essential descriptors in its proximate context, such as ``ZnO2 nanoparticles'', ``LiNbO3'', or ``half-Heuslers''. 
``Dopant'' means the elements or ions that represent a minority component, deliberately introduced impurities, or specific atomic-scale defects or carriers of electric charge like ``hole-doped'' or ``S vacancies''.
A single host can be combined with multiple dopants, through individual doping or simultaneous doping with different species, and one dopant can associate with various host materials. 
The text may contain numerous dopant-host pairings within the same sentence, and also instances of dopants and hosts that do not interact.

\begin{table}[h!]
    \centering
    \sisetup{table-format=1.3} 
    \caption{Performance comparison between LLM-NERRE, Eunomia, and Eunomia + CoV on hosts and dopants relation extraction (Case Study 1).
    Eunomia embeddings are generated using OpenAI's text-ada-002.
     Best scores for each entity are highlighted in bold text.}
    \begin{tabular}{
        l 
        l 
        S 
        S 
        S
    }
        \toprule 
        {Model} & {Entity} & {Precision (Exact Match)} & {Recall (Exact Match)} & {F1 Score (Exact Match)} \\
        \midrule 
        LLM-NERRE~\cite{dunn2022structured} & hosts & 0.892 & \textbf{0.874} & 0.883 \\
        Eunomia & hosts & 0.753 & 0.768 & 0.760 \\
        Eunomia + CoV & hosts & \textbf{0.964} & 0.853 & \textbf{0.905} \\
        \noalign{\vskip 0.5ex}
        \hdashline 
        \noalign{\vskip 0.5ex}
        LLM-NERRE~\cite{dunn2022structured} & dopants & 0.831 & 0.812 & 0.821 \\
        Eunomia & dopants & 0.859 & 0.788 & 0.822 \\
        Eunomia + CoV & dopants & \textbf{0.962} & \textbf{0.882} & \textbf{0.920} \\
    \end{tabular}
    
    \label{tab:performance_comparison_ex_1}
\end{table}

Eunomia shows an excellent performance in this task, demonstrating the effectiveness of our approach in Named Entity Recognition and Relation Extraction. 
Performance comparison between our chemist AI agent (Eunomia), and LLM-NERRE can be found in Table~\ref{tab:performance_comparison_ex_1}.
In this setting, the same above definition of hosts and dopants are passed to Eunomia via the input prompt, while LLM-NERRE is fine-tuned on 413 sentences. 
The testing set contains 77 sentences. Notably, in both tasks, Eunomia + CoV exceeds the performance of LLM-NERRE in terms of the F1 score. This clearly demonstrates the effectiveness of our approach compared to fine-tuning, which can be labor-intensive and error-prone. 
We instruct Eunomia not to make up answers, which lead to a more cautious outcome, wherein uncertain or unclear inputs yield no output.
As an example in the sentence ``An anomalous behavior of the emission properties of alkali halides doped with heavy impurities, stimulated new efforts for its interpretation, invoking delicate and sophisticated mechanisms whose interest transcends the considered specific case.'', the ground-truth host materials is ``alkali halides''. 
However, due to the nature of exact-word matching metric implemented in Ref.~\cite{dunn2022structured} a cautious agent with no predictions for the host entity will be penalized with two false negatives, one for each word in the ground-truth, leading to lower recall score.

\subsection{Case Study 2: MOF Formula and Guest Species Relation \includegraphics[height=1.45em,valign=c]{figures/medium.png}\vspace{1mm}}
\label{sec:mof-formula}
The goal of this case study is to identify MOF formula and guest species from unstructured text, as a paragraph with multiple sentences.
The MOF formula refers to the chemical formula of a MOF, which is an important piece of information for characterizing and identifying MOFs. 
The guest species, on the other hand, are chemical species that have been incorporated, stored, or adsorbed in the MOF.
These species are of interest because MOFs are often used for ion and gas separation, and the presence of specific guest species can affect the performance of the MOF. We limit our method to stand-alone Eunomia without CoV due to the complexity of defining a chemistry-informed CoV verification tool for this specific task.
It should be noted that \citet{dunn2022structured} also included results on the identification of synthesis descriptions and applications pertaining to MOFs. 
However, as the metric of exact-word matching reported in \citet{dunn2022structured} does not fairly and adequately reflect the model performance for the multi-word (> 2 words) nature of these outputs, we have limited our benchmarking to the MOF formula and guest species identification only. 

\begin{table}[h!]
    \centering
    \sisetup{table-format=1.3} 
    \caption{Performance comparison between LLM-NERRE and Eunomia on MOF formula and guest species relation extraction (Case study 2). 
    Eunomia embeddings are generated using OpenAI text-ada-002.
     Best scores for each entity are highlighted in bold text.}
    \begin{tabular}{
        l 
        l 
        S 
        S 
        S
    }
        \toprule 
        {Model} & {Entity} & {Precision (Exact Match)} & {Recall (Exact Match)} & {F1 Score (Exact Match)} \\
        \midrule 
        LLM-NERRE~\cite{dunn2022structured} & mof formula &0.409 & 0.455 & 0.424 \\
        Eunomia & mof formula & \textbf{0.623} &  \textbf{0.589} & \textbf{0.606} \\
        \noalign{\vskip 0.5ex}
        \hdashline 
        \noalign{\vskip 0.5ex}
        LLM-NERRE~\cite{dunn2022structured} & guest species & \textbf{0.588} & 0.665 & \textbf{0.606} \\
        Eunomia & guest species & 0.429 & \textbf{0.923} & 0.585 \\
    \end{tabular}
    
    \label{tab:performance_comparison_ex_2}
\end{table}

Table~\ref{tab:performance_comparison_ex_2} shows the performance comparison between Eunomia and LLM-NERRE on the MOF formula and guest species relation extraction task. While Eunomia shows a superior performance on the MOF formula compared to LLM-NERRE, the relatively low performance of both approaches is related to the nature of the exact word matching. 
Using semantic similarity would be a more appropriate indicator in this context.  
On the guest species entity, while Eunomia shows a high recall (0.923), precision is relatively poor (0.429). This can be attributed to how the exact-word matching metrics have been defined in Ref.~\cite{dunn2022structured}, where precision is majorly lowered by the presence of the extra unmatched predicted words (false positives), while recall remains high because all ground truth items were found in the predicted words.

\subsection{Case Study 3: MOF Property Relation \includegraphics[height=1.45em,valign=c]{figures/hard.png}\vspace{1mm}}
\label{sec:waterstab}
This case study aims to mimic a practical scenario of developing datasets from scientific literature, where we evaluate the agent's performance on extracting MOF's water stability. To excel in this goal, the agent must identify all MOFs mentioned within the research paper, evaluate their water stability, and support these evaluations using exact sentences derived from the document. Such tasks are inherently linked to the NLP functions of Named Entity Recognition, Co-reference Resolution, Relation Extraction, and Argument Mining. This is particularly a challenging task as researcher report the water stability in various ways, using phrases ranging from ``the material remains crystalline in humid conditions'' to ``the MOF is stable in wide range of pH'', or ``the material is not soluble in water''.

\begin{figure}[h!]
    \centering
    \includegraphics[width=0.9\textwidth]{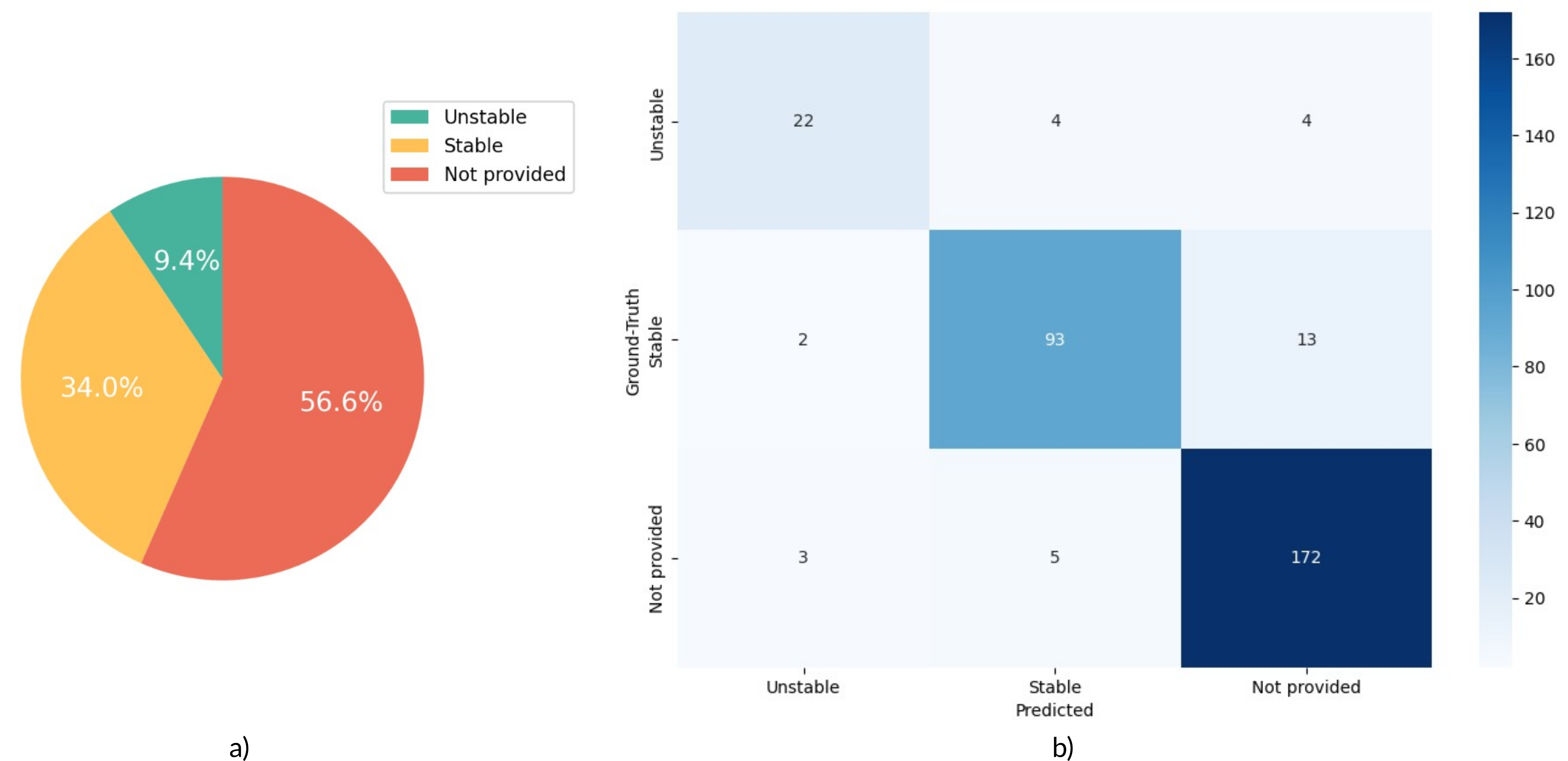}
    \caption{
    \textbf{Performance of the AI agent in information retrieval.}
    a) Class distribution for water stability in the hand-labeled ground-truth dataset of 371 MOFs based on 101 research articles. 
    b) Confusion matrix for ternary classification of water stability property with CoV tool using OpenAI text-ada-002 embeddings.
    It is apparent that our agent exercises cautious in its judgments.
    Specifically, the abundance of "Not provided" predictions, when matched against their actual ground-truth categories, suggests that the agent prefers to concede some uncertainty in instances where making an accurate prediction is not feasible, rather than incorrectly assigning samples to the "Stable" or "Unstable" categories.
    The ternary accuracy is found to be 0.91 with a yield of 86.20\%.
    \label{fig:water_stability_confusion_matrix}}
\end{figure}

For this case study, we created a hand-labeled dataset based on a selection of 101 materials research papers, which contain a selection of 371 MOFs. Three expert chemists manually read through and review each paper, pinpointing the MOFs referenced within. A portion of these articles are selected considering the original work by \citet{burtch2014water}, where they developed a dataset of MOF names and their water stability by manually reading 111 research articles. To mimic the practical data extraction scenario, in which the agent is passed many articles, many of which do not contain the desired information, we included articles with no information about water stability. 
Each MOF in our set is assigned to one of the three classes of ``Stable'', ``Unstable'', and ``Not provided''.
Figure~\ref{fig:water_stability_confusion_matrix}.a presents the distribution of the classes within this dataset.

For this case study, we have established criteria to characterize water-stable MOFs, drawing from the study by \citet{burtch2014water} and our own chemical intuition. 
A water-stable MOF should meet the following criteria:
\begin{itemize}
   \item No alteration in properties after being subjected to moisture or steam, or when    soaked or boiled in water or an aqueous solution.
    \item Preservation of its porous architecture in liquid (water) environments.
    \item Sustained crystallinity without loss of structural integrity in aqueous environments.
    \item Insoluble in water or aqueous solutions.
    \item Exhibiting a pronounced rise in its water adsorption isotherm.
\end{itemize}

These water stability guidelines are defined as rules to Eunomia within the input prompt, as well as in its equipped CoV tool.

Eunomia with CoV tool retrieve most (yield of 86.20\%) of the reported MOFs and shows an excellent performance (accuracy of 0.91) in inferring their water stability. This high yield and accuracy demonstrates the capability of our approach in extracting desired knowledge from the natural text. 
As expected, in the absence of CoV, there is a marginal decrease in accuracy to 0.86, along with a yield reduction to 82.70\%.
Taking into account the confusion matrix in Figure~\ref{fig:water_stability_confusion_matrix}.b, it is evident that our agent adopts a cautious approach in its predictions.
This is reflected in the substantial number of "Not provided" predictions which, upon comparison with the actual ground-truth class, indicates a propensity of the agent to acknowledge the insufficiency of information for making a definitive prediction, rather than mistakenly categorizing samples into the incorrect "Stable" or "Unstable" classes, and contaminating the agent's resulting dataset with unwanted noise.

\section{Discussion}
\label{sec:discuss}
We presented a high performing and robust method for extracting domain specific information from complex, unstructured text - ranging from sentences and paragraphs to extensive research articles - using AI agents. 
Scientists and researchers can use our open-source application to effortlessly develop tailored datasets for their specific areas and use them for downstream predictive tasks\cite{jablonka2023leveraging}. While currently the cost of querying large datasets may become expensive, we expect the rapid advancements in LLMs will diminish this cost.

Unlike other methods that follow a pipeline-based or end-to-end approach, our agent-based method could appeal to domain experts due to its minimal demand for programming skills, NLP and machine learning knowledge.
Users are not required to rigidly define an output schema or engage in the meticulous task of creating manual annotations for the purpose of fine-tuning. Rather, they can simply prompt the agent with more context and describe how their desired output should be formatted in natural language.
Moreover, the agent can easily be extended and equipped with other tools (e.g. Dataset Search, CSV Generator, etc.) to be adapted to other problems. For example, we showed that, by equipping the agent with the chain-of-verification tool (CoV), we can minimize Hallucinations and improve the agent's performance. Similarly, by including reasoning tools, we can ask the agent to explain its reasoning based on the provided context to develop more transparent workflows for the LLM-based methods, and reduce their known ``black-box'' nature. This, simultaneously, offers a great opportunity for human-in-the-loop oversight, especially for tasks of critical importance.

Our results reveal an important observation: while \emph{large language models are few-shot learners}~\cite{brown2020language}, AI agents with appropriate tools and instructions are capable of being zero-shot learners. This brings an excellent opportunity to boost the performance of standalone LLMs across various domain-specific tasks without having to go through labor-intensive fine-tuning processes. A future thorough and systematic analysis of prompt sensitivity can provide valuable insights into this observation.

\section{Methods}
\label{sec:methods}

\subsection{Agent Toolkits}
\label{sec:tools}
\subsubsection{Doc Search}
\label{sec:doc_search}
This tool allows for extracting relevant knowledge materials properties from text, ranging from a single sentence and paragraph to a scientific research paper.
The research papers are obtained from various chemistry journals including Royal Society of Chemistry (RSC), American Chemical Society (ACS), Elsevier, Inorganic Chemistry, Structural Chemistry, Coordination Chemistry, Wiley, and Crystallographic Communications as a PDF or in XML format (the XML files are obtained through a legal agreement between University of Toronto and ACS).
Inspired by the paper-qa Python package~\href{https://github.com/whitead/paper-qa}{(https://github.com/whitead/paper-qa)}, this tool aims at obtaining the most relevant context (sentences) from the papers to a given input query.
This involves embedding the paper and queries into numerical vectors and identifying top $k$ passages within the document that either mention or can somehow imply the property of interest for a MOF.
$k$ is set to 9 in our case studies, and is dynamically adjusted depending on the length of the paper to avoid OpenAI's token limitation error.
We use OpenAI's text-ada-002 embeddings~\cite{greene2022new} to represent texts as high dimensional vectors, which are stored as a vector database using FAISS~\cite{johnson2019billion}.
Note that the choice of embedding is another hyperparameter that can be changed in future studies. 
For benchmarking purposes, we have also conducted all case studies with the newly released Cohere embed-english-v3.0 embeddings (see Supporting Information).

The semantic similarity search is ranked using Maximum Marginal Relevance (MMR)~\cite{carbonell1998use} based on cosine similarity, defined as,
\begin{equation}
\text{MMR} = \arg\max_{d_i \in R \setminus S} \left[ \lambda \cdot \cos(d_i, q) - (1 - \lambda) \cdot \max_{d_j \in S} \cos(d_i, d_j) \right]
\label{eq:MMR}
\end{equation}
where $d_i$ is a document from the set of retrieved 
documents $R$, $S$ is the set of already selected documents, $q$ is the query.
$\lambda$ is a parameter between 0 and 1 that balances the trade-off between relevance (to the query) and diversity (or novelty with respect to already selected documents).
In this work, we use the default value of 0.5.
The idea behind MMR is to retrieve or select documents that are not just relevant to the query (or topic of interest), but are also diverse among themselves, thus minimizing redundancy.
This tool provides the exact in-context sentence from the paper that provides some reasoning for the agent's choice, allowing for a more methodical evaluation of the agent's decision-making process, and reducing the likelihood of hallucinations or fabrications with the human-in-loop verification of the resulting datasets.
It is important to note that in some unsuccessful experiments, we observed that the AI agent repeatedly referred back to the document, even after pinpointing the correct answer. 
Although this minor issue remained unresolved, we introduced an iteration limit for the agent to avoid unnecessary model running costs.

\subsubsection{Chain-of-Verification}
\label{sec:CoV}
Inspired by the Chain-of-Verification (CoV)~\cite{dhuliawala2023chain} methodology, this tool entails the following steps: initially, the agent provides a preliminary reply, which is followed by iterative verification queries to authenticate the initial draft.
The agent independently responds to these queries to ensure the answers remain impartial and unaffected by other responses, and finally it produces its conclusive, verified response.
Our implementation of CoV stands apart from the method described in \citet{dhuliawala2023chain}, specifically in how the verification queries are generated. 
While in the \citet{dhuliawala2023chain}'s approach, the LLM produces task-specific queries, our method allows for user customization. 
This adaptability not only enables broader, more tailored domain-specific fact-checking across various tasks, but also opens up opportunities for human-in-the-loop verification, enhancing the accuracy and relevance of the results.
This tool substantially boosts agent efficacy and mitigates the likelihood of hallucinations, especially in the events of completing complex tasks (see Figure~\ref{fig:CoV} for more details).
It is important to note that for unknown reasons, we have observed that the CoV tool usage was skipped by the agent on a few occasional instances.

\subsubsection{Dataset Search}
This tool allows for obtaining the chemical structure of MOFs from publically available datasets, including the Materials Projects~\cite{jain2013commentary}, Crystallography Open Database (COD)~\cite{merkys2023graph, vaitkus2021validation, quiros2018using,merkys2016cod,gravzulis2015computing, gravzulis2012crystallography, gravzulis2009crystallography, downs2003american}, Cambridge Structural Database (CSD)~\cite{groom2016cambridge}, and QMOF~\cite{rosen2021machine, rosen2022high}.
\subsubsection{CSV Generator}
This tool stores the output of the agent into a CSV or JSON file.

\subsection{Evaluation Metrics}
\label{sec:eval-metrics}
Multiple metrics have been defined to assess agent's performance across different case studies. Precision, recall and F1 score are defined as,

\begin{equation}
\text{Precision} = \frac{TP}{TP + FP},
\label{eq:precision}
\end{equation}

\begin{equation}
\text{Recall} = \frac{TP}{TP + FN},
\label{eq:recall}
\end{equation}

\begin{equation}
\text{F1 Score} = 2 \times \frac{\text{Precision} \times \text{Recall}}{\text{Precision} + \text{Recall}},
\label{eq:f1score}
\end{equation}

where, $TP$ represents true positives, $FP$ stands for false positives, and $FN$ denotes false negatives. 
Precision measures the accuracy of the positive predictions, recall measures the fraction of actual positives that were correctly identified, and the F1 score is the harmonic mean of precision and recall. Binary classification accuracy is defined as,

\begin{equation}
\text{Binary accuracy} = \frac{TP + TN}{N},
\label{eq:acc_t}
\end{equation}

In Case studies 1 and 2 (Sections \ref{sec:dopant} and \ref{sec:mof-formula}), the evaluation metrics used are precisely those defined in the work of ~\cite{dunn2022structured}.
In specific, they assessed named entity relationships on a word-to-word matching basis by initially decomposing an entity \(E\) into a collection of \(k\) words separated by whitespace, denoted as \(E = \{w_1, w_2, w_3, \ldots, w_k\}\). For evaluating entities in Named Entity Recognition exclusively, they enumerated the words that are identical in both the true entity set \(E_{\text{true}}\) and the test entity set \(E_{\text{test}}\) as true positives (\(E_{\text{true}} \cap E_{\text{test}}\)), and the distinct elements in each set as false positives (\(E_{\text{test}} \setminus E_{\text{true}}\)) or false negatives (\(E_{\text{true}} \setminus E_{\text{test}}\)). 
 For instance, if the true entity is ``Bi$_2$Te$_3$ thin film'' and the predicted entity is ``Bi$_2$Te$_3$ film sample'', they noted two true positive matches (``Bi$_2$Te$_3$'', ``film''), one false negative (``thin''), and one false positive (``sample''). 
 An exceptional case arises for formula-type entities critical to material identification, whereby \(E_{\text{test}}\) must encompass all \(w_i\) interpreted as stoichiometries to consider any \(w_i \in E_{\text{test}}\) as correct. 
 For example, with ``Bi$_2$Te$_3$ thin film'' as \(E_{\text{true}}\) and ``thin film'' as \(E_{\text{test}}\), three false negatives would be registered.
 For more details on the scoring metrics and the case studies, readers are encouraged to refer to Ref.~\cite{dunn2022structured}.

For our last case study in Section~\ref{sec:waterstab} (predicting water stability of MOFs), the ternary accuracy is defined as,
\begin{equation}
\text{Ternary accuracy} = \frac{TP_S + TP_U + TP_{NP}}{N},
\label{eq:acc}
\end{equation}
where $N$ shows the total number of predictions and $S$, $U$, $NP$ denote the three classes ``Stable'', ``Unstable'', and ~``Not provided'', respectively.
$TP_i$ shows then number of instances correctly predicted as class $i$. 
Additionally, we evaluate the information recovery capabilities of the agent by defining yield as
\begin{equation}
\text{Yield} = \frac{N}{N_{GT}},
\label{eq:yield}
\end{equation}
where $N_{GT}$ is the \emph{ground-truth} number of MOFs mentioned in the research papers, regardless whether the paper discusses water stability or not.

\section*{Acknowledgements}
Research reported in this work was supported by the Acceleration Consortium at the University of Toronto. SMM acknowledges the support by the Natural Sciences and Engineering Research Council of Canada (NSERC) under grant number RGPIN-2023-04232. The authors thank Alexander Dunn and Andrew S. Rosen for their assistance on the implementation of LLM-NERRE.
The authors also thank Haoning Yuan for assisting in manual curation of the water stability dataset used in Section~\ref{sec:waterstab}.

\section*{Data and Code Availability}
All data (including the dataset used for case study 3) and code used to produce results in this study are publically available in the following GitHub repository:
\href{https://github.com/AI4ChemS/Eunomia}{\color{blue}{{https://github.com/AI4ChemS/Eunomia}}}.
The methodology in this work is also developed as an open-source application on \href{https://eunomia.streamlit.app/}{\color{blue}{{https://eunomia.streamlit.app}}}.


\bibliographystyle{unsrtnat}
\bibliography{bibliography}

\end{document}


\title{Supporting Information for Agent-based Learning of Materials datasets from scientific literature}

\author[1,2]{Mehrad Ansari}
\author[ \space ,1,2]{Seyed Mohamad Moosavi\thanks{Corresponding author: mohamad.moosavi@utoronto.ca}}

\affil[1]{Acceleration Consortium, University of Toronto, Toronto, Ontario M5S 3E5, Canada}
\affil[2]{ Department of Chemical Engineering \& Applied Chemistry, University of Toronto, Toronto, Ontario M5S 3E5, Canada}

\maketitle

    \centering

\section{Supporting Figures}

\begin{figure}[h!]
    \centering
    \includegraphics[width=0.8\textwidth]{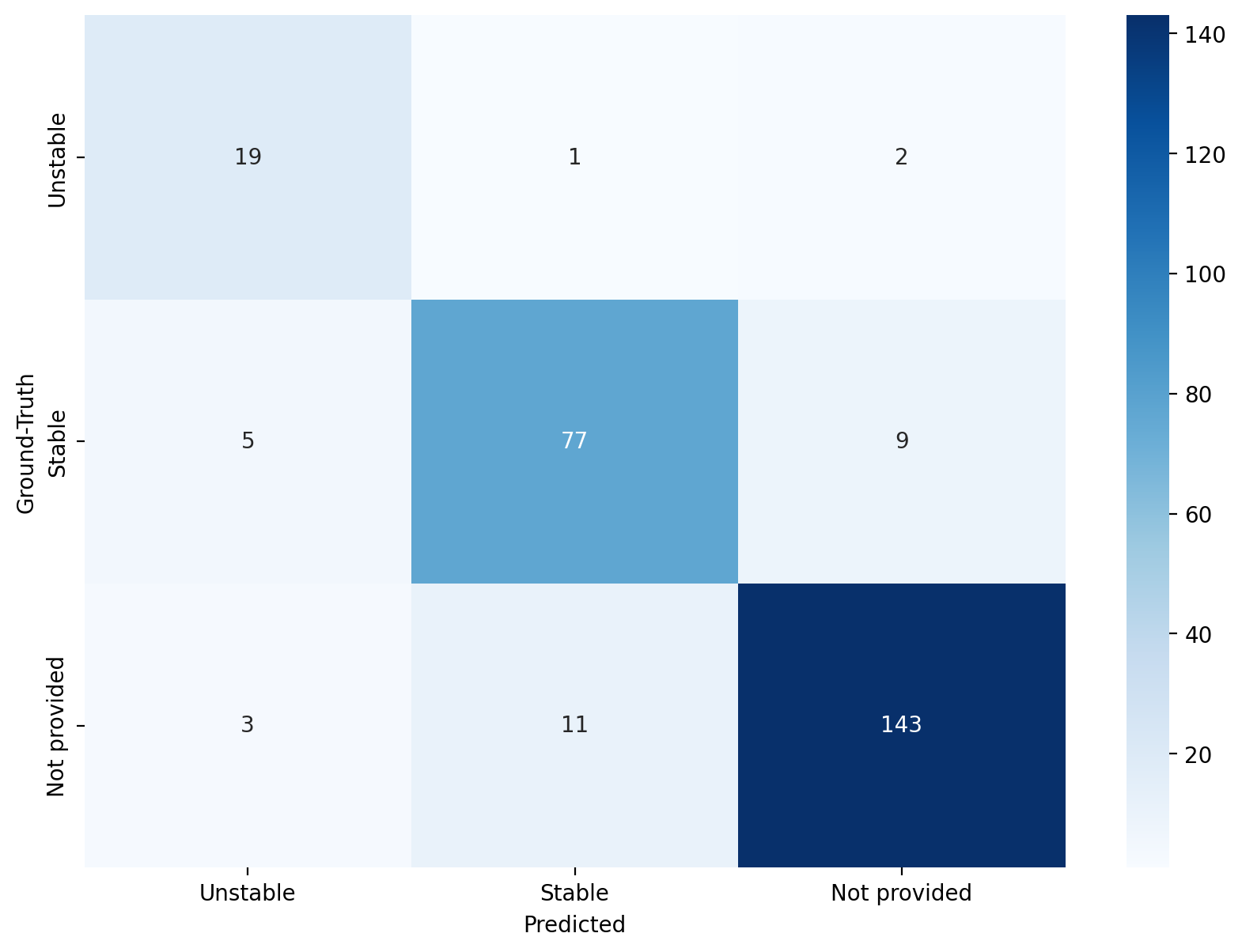}
    \caption{Confusion matrix for ternary classification of water stability property in Case Study 3 using Cohere embed-english-v3.0 embeddings.
    It is apparent that our agent exercises cautious in its judgments.
    Specifically, the abundance of "Not provided" predictions, when matched against their actual ground-truth categories, suggests that the agent prefers to concede some uncertainty in instances where making an accurate prediction is not feasible, rather than incorrectly assigning samples to the "Stable" or "Unstable" categories.
    The ternary accuracy is found to be 0.88 with a yield of 72.96\%, which is lower than performance coming from OpenAI's text-ada-002 embeddings (see Figure 4b in the main article).
    \label{fig:water_stability_confusion_matrix}}
\end{figure}
    
\section{Supporting Tables}

\begin{table}[h!]
    \centering
    \sisetup{table-format=1.3} 
    \caption{Performance comparison between LLM-NERRE, Eunomia, and Eunomia + CoV on hosts and dopants relation extraction (Case Study 1).
    Eunomia embeddings are generated using Cohere embed-english-v3.0.}
    \begin{tabular}{
        l 
        l 
        S 
        S 
        S
    }
        \toprule 
        {Model} & {Entity} & {Precision (Exact Match)} & {Recall (Exact Match)} & {F1 Score (Exact Match)} \\
        \midrule 
        LLM-NERRE & hosts & 0.892 & 0.874 & 0.883 \\
        Eunomia & hosts & 0.710 & 0.747 & 0.728 \\
        Eunomia + CoV & hosts & 0.793 & 0.726 & 0.758 \\
        \noalign{\vskip 0.5ex}
        \hdashline 
        \noalign{\vskip 0.5ex}
        LLM-NERRE & dopants & 0.831 & 0.812 & 0.821 \\
        Eunomia & dopants & 0.782 & 0.800 & 0.791 \\
        Eunomia + CoV & dopants & 0.863& 0.812 & 0.836 \\
    \end{tabular}
    
    \label{tab:SI_performance_comparison_ex_1}
\end{table}

\begin{table}[h!]
    \centering
    \sisetup{table-format=1.3} 
    \caption{Performance comparison between LLM-NERRE and Eunomia on MOF formula and guest species relation extraction (Case study 2). 
    Eunomia embeddings are generated using Cohere embed-english-v3.0.}
    \begin{tabular}{
        l 
        l 
        S 
        S 
        S
    }
        \toprule 
        {Model} & {Entity} & {Precision (Exact Match)} & {Recall (Exact Match)} & {F1 Score (Exact Match)} \\
        \midrule 
        LLM-NERRE & mof formula &0.409 & 0.455 & 0.424 \\
        Eunomia & mof formula & \textbf{0.615} &  \textbf{0.571} & \textbf{0.593} \\
        \noalign{\vskip 0.5ex}
        \hdashline 
        \noalign{\vskip 0.5ex}
        LLM-NERRE & guest species & \textbf{0.588} & 0.665 & \textbf{0.606} \\
        Eunomia & guest species & 0.213 & \textbf{0.769} & 0.333 \\
    \end{tabular}
    
    \label{tab:performance_comparison_ex_2}
\end{table}
